\documentclass{article}
\pdfoutput=1

\usepackage{graphicx} % more modern
\usepackage{subfigure} 

\usepackage{natbib}

\usepackage{algorithm}
\usepackage{algorithmic}

\usepackage{hyperref}
\usepackage{xcolor}
\usepackage[accepted]{icml2012}

\icmltitlerunning{Unachievable Region in Precision-Recall Space}

\usepackage{amsmath}
\usepackage{amsthm}
\usepackage{verbatim}

\usepackage{xspace}
\usepackage{comment}

\theoremstyle{plain}
\newtheorem{theorem}{Theorem}
\newtheorem{corollary}[theorem]{Corollary}

\theoremstyle{definition}
\newtheorem{definition}{Definition}

\newcommand{\maketable}[4]{ 
  \begin{table}
    \caption{#2}
    \label{#1} 
    \vskip 0.15in
    \begin{center}
      \begin{small}
        \begin{sc}
          \begin{tabular}{#3}
            #4
          \end{tabular}
        \end{sc}
      \end{small}
    \end{center}
    \vskip -0.15in
  \end{table}
}

\newcommand{\makefigure}[3]{
  \begin{figure}
    \vskip 0.1in
    \begin{center}
      \centerline{\includegraphics[width=\columnwidth]{#3}}
      \caption{#2}
      \label{#1}
    \end{center}
    \vskip -0.3in      
  \end{figure}
}

\newcommand{\myeqref}[1]{
  Eq.~\eqref{#1}\xspace
}

\newcommand{\aucpr}{
  \ensuremath{\mathrm{AUCPR}}\xspace
}

\newcommand{\aucprmin}{
  \ensuremath{\mathrm{AUCPR}_{\mathrm{MIN}}}\xspace
}

\newcommand{\aucprmax}{
  \ensuremath{\mathrm{AUCPR}_{\mathrm{MAX}}}\xspace
}

\newcommand{\aucnpr}{
  \ensuremath{\mathrm{AUCNPR}}\xspace
}

\newcommand{\tp}{\ensuremath{\mathit{tp}}\xspace}
\newcommand{\fp}{\ensuremath{\mathit{fp}}\xspace}
\newcommand{\fn}{\ensuremath{\mathit{fn}}\xspace}
\newcommand{\tn}{\ensuremath{\mathit{tn}}\xspace}
\newcommand{\numpos}{\ensuremath{\mathit{pos}}\xspace}
\newcommand{\numneg}{\ensuremath{\mathit{neg}}\xspace}

\usepackage{tabularx}
\begin{document} 

\twocolumn[
\icmltitle{Unachievable Region in Precision-Recall Space\\and Its Effect on Empirical Evaluation}

% unless you've provided the [accepted] option to the icml2012
%\icmlauthor{Your Name}{email@yourdomain.edu}
%\icmladdress{Your Fantastic Institute,
%            314159 Pi St., Palo Alto, CA 94306 USA}
%\icmlauthor{Your CoAuthor's Name}{email@coauthordomain.edu}
%\icmladdress{Their Fantastic Institute,
%            27182 Exp St., Toronto, ON M6H 2T1 CANADA}
\icmlauthor{Kendrick Boyd}{boyd@cs.wisc.edu}
\icmladdress{University of Wisconsin -- Madison, 1300 University Avenue, Madison, WI 53706 USA}
\icmlauthor{V\'{\i}tor Santos Costa}{vsc@dcc.fc.up.pt}
\icmladdress{CRACS INESC-TEC \& FCUP, Rua do Campo Alegre, 1021/1055, 4169 - 007 PORTO, Portugal}
\icmlauthor{Jesse Davis}{jesse.davis@cs.kuleuven.be}
\icmladdress{KU Leuven, Celestijnenlaan 200a, Heverlee 3001, Belgium}
\icmlauthor{C. David Page}{page@biostat.wisc.edu}
\icmladdress{University of Wisconsin -- Madison, 1300 University Avenue, Madison, WI 53706 USA}


\icmlkeywords{}

\vskip 0.3in
]

\begin{abstract}
Precision-recall (PR) curves and the areas under them are widely used to summarize machine learning results, especially for data sets exhibiting class skew.
They are often used analogously to ROC curves and the area under ROC curves.
It is known that PR curves vary as class skew changes.
What was not recognized before this paper is that there is a region of PR space that is completely unachievable, and the size of this region depends only on the skew.
This paper precisely characterizes the size of that region and discusses its implications for empirical evaluation methodology in machine learning.

\end{abstract}

\section{Introduction}

Precision-recall (PR) curves are a common way to evaluate the performance of a
machine learning algorithm. PR curves illustrate the tradeoff between
the proportion of positively labeled examples that are truly positive
(precision) as a function of the proportion of correctly classified
positives (recall).  In particular, PR analysis is preferred to ROC
analysis when there is a large skew in the class distribution. In this
situation, even a relatively low false positive {\em rate} can produce
a large number of false positives and hence a low
precision~\cite{davis2006}. Many applications are
characterized by a large skew in the class distribution. In
information retrieval (IR), only a few documents are relevant to a
given query.  In medical diagnoses, only a small proportion of the
population has a specific disease at any given time.  In relational
learning, only a small fraction of the possible groundings of a
relation are true in a database.

The area under the precision-recall curve (AUCPR) often serves as a
summary statistic when comparing the performance of different
algorithms.  For example, IR systems are frequently judged by their mean
average precision, or MAP (not to be confused with the same acronym for
``maximum a posteriori''), which is an approximation of the mean AUCPR
over the queries \cite{manning2008}.  Similarly, AUCPR often serves as an evaluation criteria for statistical relational learning
(SRL) \cite{kok2010,davis2005,ilya2010,mihalkova2007} and information
extraction (IE) \cite{ling2010,goadrich2006}.  Additionally, some algorithms,
such as SVM-MAP \cite{yue2007} and SAYU \cite{davis2005}, explicitly optimize the AUCPR of the learned model.

There is a growing body of work that analyzes the properties of PR
curves \cite{davis2006,clemencon2009}.  Still, PR curves and AUCPR are
frequently treated as a simple substitute in skewed domains for ROC
curves and area under the ROC curve (AUCROC), despite the known
differences between PR and ROC curves.  These differences include that
for a given ROC curve the corresponding PR curve varies with class
skew~\cite{davis2006}.  A related, but previously unrecognized,
distinction between the two types of curves is that, while any point
in ROC space is achievable, not every point in PR space is achievable.
That is, for a given data set it is possible to construct a confusion
matrix that corresponds to any (false positive rate, true positive rate)
pair, but it is \emph{not possible} to do this for every (recall, precision) pair.\footnote{To be strictly true in ROC space, fractional
counts for $tp,fp,fn,tn$ must be allowed. The fractional
counts can be considered integer counts in an expanded data set.}

We show that this distinction between ROC space and PR space has major
implications for the use of PR curves and AUCPR in machine learning.
The foremost is that the unachievable points define a minimum PR
curve.  The area under the minimum PR curve constitutes a portion of
AUCPR that any algorithm, no matter how poor, is guaranteed to obtain
``for free.''  Figure~\ref{fig:prcurve:example} illustrates the
phenomenon.  Interestingly, we prove that the size of the unachievable
region is
\emph{only a function of class skew} and has a simple, closed form.

The unachievable region can influence algorithm evaluation and even
behavior in many ways.  Even for evaluations using F1 score, which only
consider a single point in PR space, the unachievable region has
subtle implications.  When averaging AUCPR over multiple tasks (e.g.,
SRL target predicates or IR queries), the area under the minimum PR
curve alone for a non-skewed task may outweigh the total area for all
other tasks.  A similar effect can occur when the folds used for
cross-validation do not have the same skew. Downsampling that changes
the skew will also change the minimum PR curve.  In algorithms that
explicitly optimize AUCPR or MAP during training, algorithm behavior
can change substantially with a change in skew.  These undesirable
effects of the unachievable region can be at least partially offset
with straightforward modifications to AUCPR, which we describe.

\makefigure{fig:prcurve:example}{Minimum PR curve and random guessing curve at a skew of 1 positive for every 2 negative examples.}{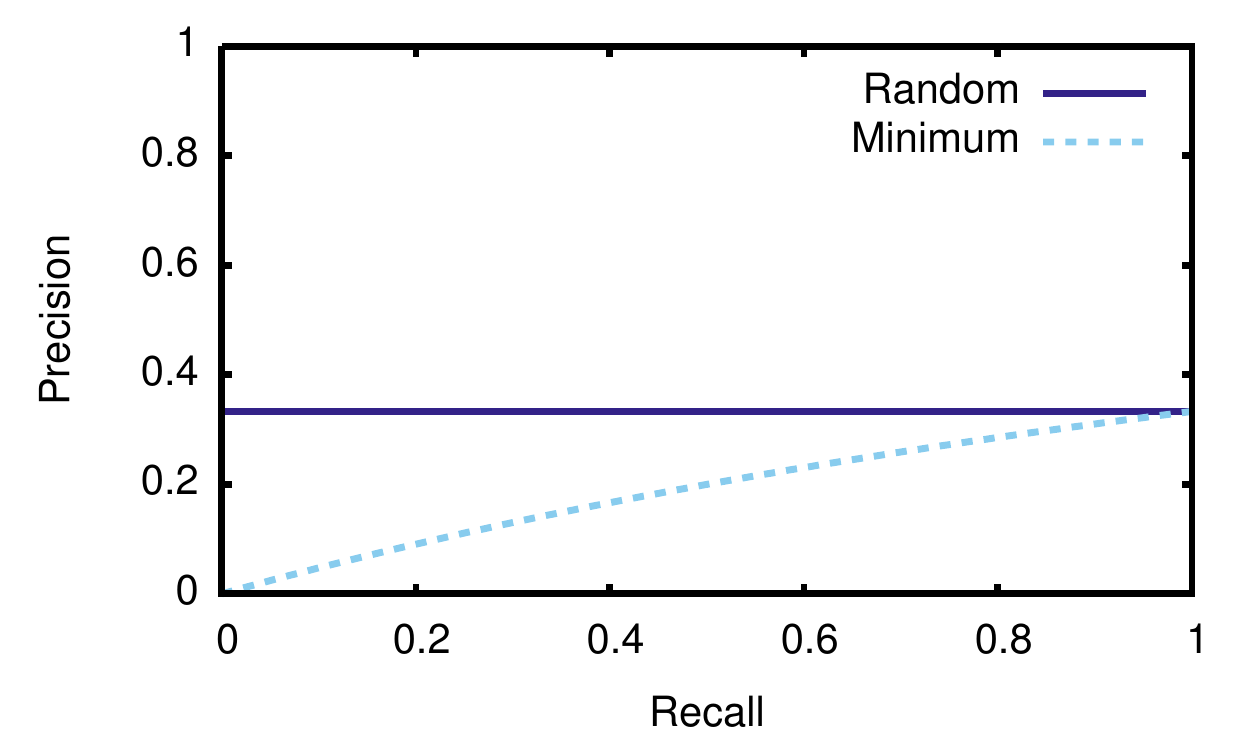}

\section{Achievable and Unachievable Points in PR Space}

We first precisely define the notion of an achievable point in PR
space. Then we provide an intuitive example to illustrate the concept of
an unachievable point.  Finally, in Theorems~\ref{thm:constraint}~and~\ref{thm:minarea} we present our central theoretical contributions that formalize the notion of the unachievable region in PR space.

We assume familiarity with precision, recall, and confusion matrices
(see Davis and Goadrich \yrcite{davis2006} for an
overview).  We use $p$ for precision, $r$ for recall, and
$\tp,\fp,\fn,\tn$ for the number of true positives, false positives, false
negatives, and true negatives, respectively.

Consider a data set $D$ with $n = \numpos + \numneg$ examples, where $\numpos$ is
the number of positive examples and $\numneg$ is the number of negative
examples.  A \emph{valid} confusion matrix for $D$ is a tuple
$(\tp,\fp,\fn,\tn)$ such that $\tp, \fp, \fn, \tn \geq 0$, $\tp + \fn = \numpos$ and $\fp + \tn = \numneg$.
We use  $\pi = \frac{\numpos}{n}$, the proportion of examples that are positive, to quantify the skew of $D$. Following convention, highly skewed refers to $\pi$ near 0 and non- or less skewed to $\pi$ near 0.5.

\begin{definition} For a data set $D$, an \emph{achievable point} in PR
space is a point $(r,p)$ such that there exists a valid confusion matrix with recall $r$ and
precision $p$.
\end{definition}

\subsection{Unachievable Points in PR Space}

One can easily show that, like in ROC space, each valid confusion
matrix, where $\tp > 0$, defines a single and unique point in PR space.
In PR space, both recall and precision depend on the $\tp$ cell of the
confusion matrix, in contrast to the true positive rate and false
positive rate used in ROC space.  This dependence, together with the
fact that a specific data set contains a fixed number of negative and
positive examples, imposes limitations on what precisions are possible
for a particular recall.

To illustrate this effect, consider a data
set with $\numpos=100$ and $\numneg=200$.
Table~\ref{tbl:matrix_valid} shows a valid confusion matrix with $r=0.2$ and $p=0.2$.  Consider holding precision constant while increasing recall.
Obtaining $r=0.4$ is possible with $\tp=40$ and $\fn=60$. Notice that keeping $p=0.2$ requires increasing $fp$ from 80 to 160.
With a fixed number of negative examples in the data set, increases in $fp$ cannot continue indefinitely.
For this data set, $r=0.5$ with $p=0.2$ is possible by using all negatives as false positives (so $\tn=0$).
However, maintaining $p=0.2$ for any $r>0.5$ is impossible.
 Table~\ref{tbl:matrix_invalid} illustrates an attempted confusion matrix with $r=0.6$ and $p=0.2$.
Achieving $p=0.2$ at this recall requires $\fp > \numneg$.
This forces $\tn < 0$ and makes the confusion matrix invalid.

\begin{table}  
  \caption{\subref{tbl:matrix_valid} Valid confusion matrix with $r=0.2$ and $p=0.2$ and \subref{tbl:matrix_invalid} invalid confusion matrix attempting to obtain $r=0.6$ and $p=0.2$.}
  \label{tbl:matrices}
  \vskip 0.1in
  \begin{center}
    \subfigure[Valid]{ \label{tbl:matrix_valid}
    \begin{small}\begin{sc}
      \begin{tabular}{|l|c|c|}
      \hline
      & \multicolumn{2}{c|}{Actual} \\
      Label & Pos & Neg \\ \hline
      Pos & 20 & 80 \\ \hline
      Neg & 80 & 120 \\ \hline
      Total & 100 & 200 \\ \hline
      \end{tabular}
    \end{sc}\end{small}
    }    
    \subfigure[Invalid]{ \label{tbl:matrix_invalid}
    \begin{small}\begin{sc}
      \begin{tabular}{|l|c|c|}
      \hline
      & \multicolumn{2}{c|}{Actual} \\
      Label & Pos & Neg \\ \hline
      Pos & 60 & 240 \\ \hline
      Neg & 40 & \textbf{-40} \\ \hline
      Total & 100 & 200 \\ \hline
      \end{tabular}
      \end{sc}\end{small}
    }
  \end{center}
  \vskip -0.15in
\end{table}

The following theorem formalizes this restriction on achievable points
in PR space.

\begin{samepage}
  \begin{theorem} \label{thm:constraint} Precision ($p$) and recall
($r$) must satisfy,
    \begin{equation} \label{eq:constraint} p \geq \frac{\pi r}{1-\pi
+ \pi r}
    \end{equation} where $\pi$ is the skew.
  \end{theorem}
\end{samepage}

\begin{proof}
  Starting from the definition of precision,
  \begin{equation*}
    p = \frac{\tp}{\tp + \fp} \geq \frac{\tp}{\tp + (1-\pi)n}
  \end{equation*}
  since the false positives cannot be greater than the number of negatives.
  $\tp = r \pi n$ from the definition of recall, and we can reasonably assume the data set is non-empty ($n > 0$) so the $n$s cancel out. Thus
  \begin{equation*}
    p \geq \frac{r \pi n}{r \pi n + (1-\pi)n} = \frac{r \pi}{r \pi + 1 - \pi} \qedhere
  \end{equation*}
\end{proof}

If a point in PR space satisfies\myeqref{eq:constraint}, we say it is
\emph{achievable}.  Note that a point's
achievability depends solely on the skew and not on a data set's size.
Thus we often refer to achievability in terms of the skew and not in reference to any particular data set.

\subsection{Unachievable Region in PR Space} 
Theorem~\ref{thm:constraint} gives a constraint that each achievable
point in PR space must satisfy.  For a given skew, there are many
points that are unachievable, and we refer to this collection of
points as the \emph{unachievable region} of PR space. This subsection
studies the properties of the unachievable region.

\myeqref{eq:constraint} makes no assumptions about
a model's performance. Consider a model that gives the worst possible
ranking where every negative example is ranked ahead of every positive
example. Building a PR curve based on this ranking means placing one
PR point at $(0,0)$ and a second PR point at $(1,\frac{\numpos}{n})$.
Davis and Goadrich \yrcite{davis2006} provide the correct method for
interpolating between points in PR space; interpolation is non-linear in PR space
but is linear between the corresponding points in ROC
space. Interpolating between the two known points gives intermediate
points with recall of $r_i = \frac{i}{n}$ and precision of $p_i =
\frac{\pi r_i}{(1-\pi) + r_i \pi}$, for $0 \leq i \leq pos$.  This is the equality case from Theorem~\ref{thm:constraint}, so\myeqref{eq:constraint} is a
tight lower bound on precision.  We call the curve produced by this
ranking the \emph{minimum PR curve} because it lies on the boundary
between the achievable and unachievable regions of PR space. For a
given skew, all achievable points are on or above the minimum PR curve.

\makefigure{fig:prcurve:skews}{Minimum PR curves for several values of $\pi$.}{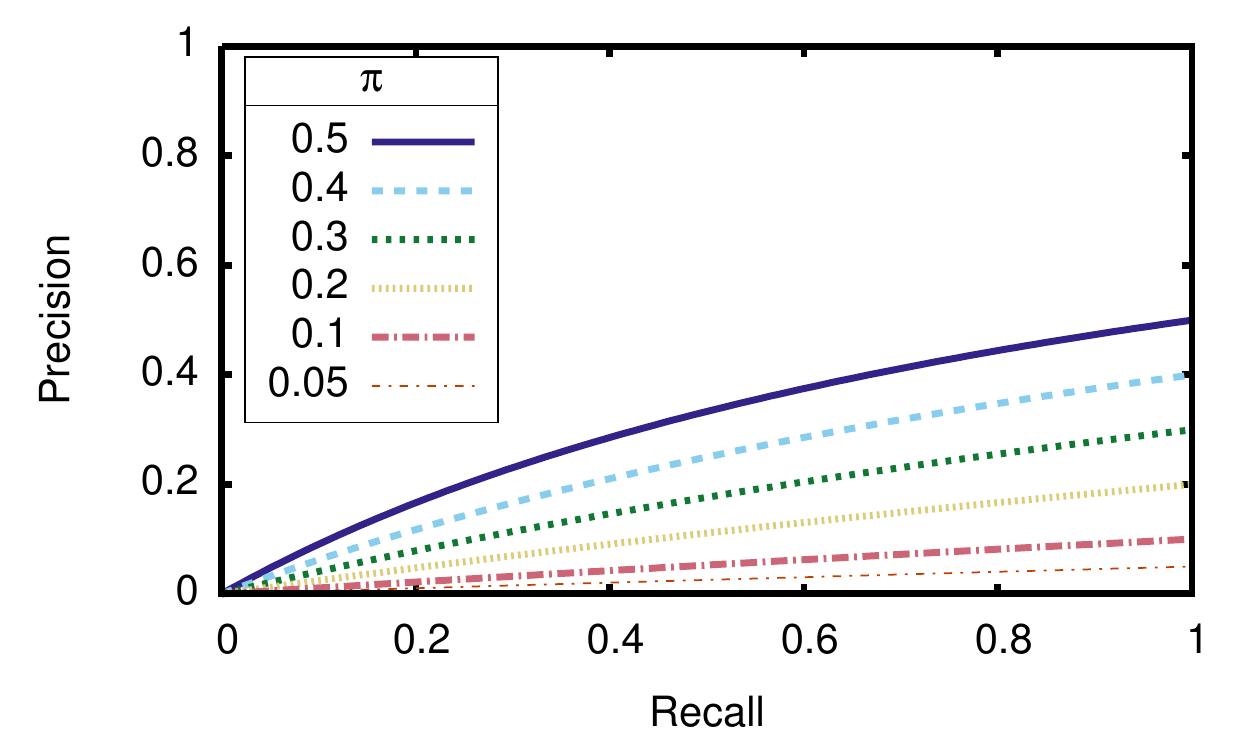}

The minimum PR curve has an interesting implication for AUCPR and average precision.  Any model must produce a PR curve that lies above the minimum PR curve.
Thus, the AUCPR score includes the size
of the unachievable region ``for free.''  In the following theorem, we
provide a closed form solution for calculating the area of the unachievable
region.
\begin{samepage}
\begin{theorem} \label{thm:minarea} The area of the unachievable
region in PR space and the minimum\aucpr, for skew $\pi$, is
  \begin{equation} \label{eq:minarea}
   \aucprmin = 1 + \frac{(1 - \pi)\ln (1-\pi)}{\pi}
  \end{equation}
\end{theorem}
\end{samepage}
\begin{proof} Since\myeqref{eq:constraint} gives a lower bound for the
precision at a particular recall, the unachievable
area is the area below the curve $f(r) = \frac{r \pi}{1-\pi + r \pi}$.

  \begin{align*} 
  \aucprmin &= \int_0^1 \frac{r \pi}{1-\pi + r\pi}\,\mathrm{d}r \\
   &= \left. \frac{r \pi + (\pi - 1)\ln(\pi(r-1) + 1)}{\pi} \right |_{r=0}^{r=1} \\
   &= \frac{1}{\pi} (\pi + (\pi - 1)(\ln(1) - \ln (1 - \pi))) \\
   &= 1 + \frac{(1 - \pi)\ln (1-\pi)}{\pi} \qedhere
  \end{align*}
\end{proof}

See Figure~\ref{fig:minareas_line} for\aucprmin at different skews.

Similar to AUCPR,\myeqref{eq:constraint} also defines a minimum for average precision (AP).
Average precision is the mean precision after correctly labeling each positive example in the ranking, so the minimum takes the form of a discrete summation. Unlike AUCPR, which is calculated from interpolated curves, the minimum AP depends on the number of positive examples because that controls the number of terms in the summation.

\begin{samepage}
\begin{theorem} 
The minimum $\mathrm{AP}$, for $\numpos$ and $\numneg$ positive and negative examples, respectively, is
\begin{equation*}
  \mathrm{AP}_{\mathrm{MIN}} = \frac{1}{\numpos}\sum_{i=1}^{\numpos} \frac{i}{i+\numneg}
\end{equation*}
\end{theorem}
\end{samepage}
\begin{proof}
  \begin{align*}
  \mathrm{AP}_{\mathrm{MIN}} &= \frac{1}{\numpos} \sum_{i=1}^{\numpos} \frac{\frac{\pi i}{\numpos}}{1 - \pi + \frac{\pi i}{\numpos}} \\
  &= \frac{1}{\numpos} \sum_{i=1}^{\numpos} \frac{\frac{\numpos i}{(\numpos + \numneg)\numpos}}{1 + \frac{\numpos}{\numpos+\numneg}(\frac{i}{\numpos} - 1)} \\
  &= \frac{1}{\numpos} \sum_{i=1}^{\numpos} \frac{\frac{i}{\numpos+\numneg}}{\frac{i+\numneg}{\numpos + \numneg}} 
  = \frac{1}{\numpos} \sum_{i=1}^{\numpos} \frac{i}{i + \numneg} \qedhere
  \end{align*}
\end{proof}
This precisely captures the natural intuition that the worst AP involves labeling all negatives examples as positive before starting to label the positives.

The existence of the minimum AUCPR and minimum AP can affect the qualitative
interpretation of a model's performance.  For example, changing the
skew of a data set from 0.01 to 0.5 (e.g., by subsampling the negative
examples \cite{natarajan2011,ilya2010}) increases the minimum AUCPR by approximately 0.3.
This leads to an automatic jump of 0.3 in AUCPR simply by changing the data set and with absolutely no change to the learning algorithm.

Since the majority of the unachievable region is at higher recalls, the effect of\aucprmin becomes more pronounced when restricting the area calculation to high levels of recall.  Calculating AUCPR for recalls
above a threshold is frequently done due to the high variance of
precision at low recall or because the learning problem requires high
recall solutions (e.g., medical domains such as breast cancer risk prediction).
Corollary~\ref{thm:minarearestricted} gives the formula for computing\aucprmin when the area is calculated over a restricted range of recalls. See Figure~\ref{fig:minareas_line} for minimum AUCPR when calculating area over restricted recall.

\makefigure{fig:minareas_line}{Minimum AUCPR versus $\pi$ for area calculated over recall in [0,1] (entire PR curve), [0.5,1], and [0.8,1].}{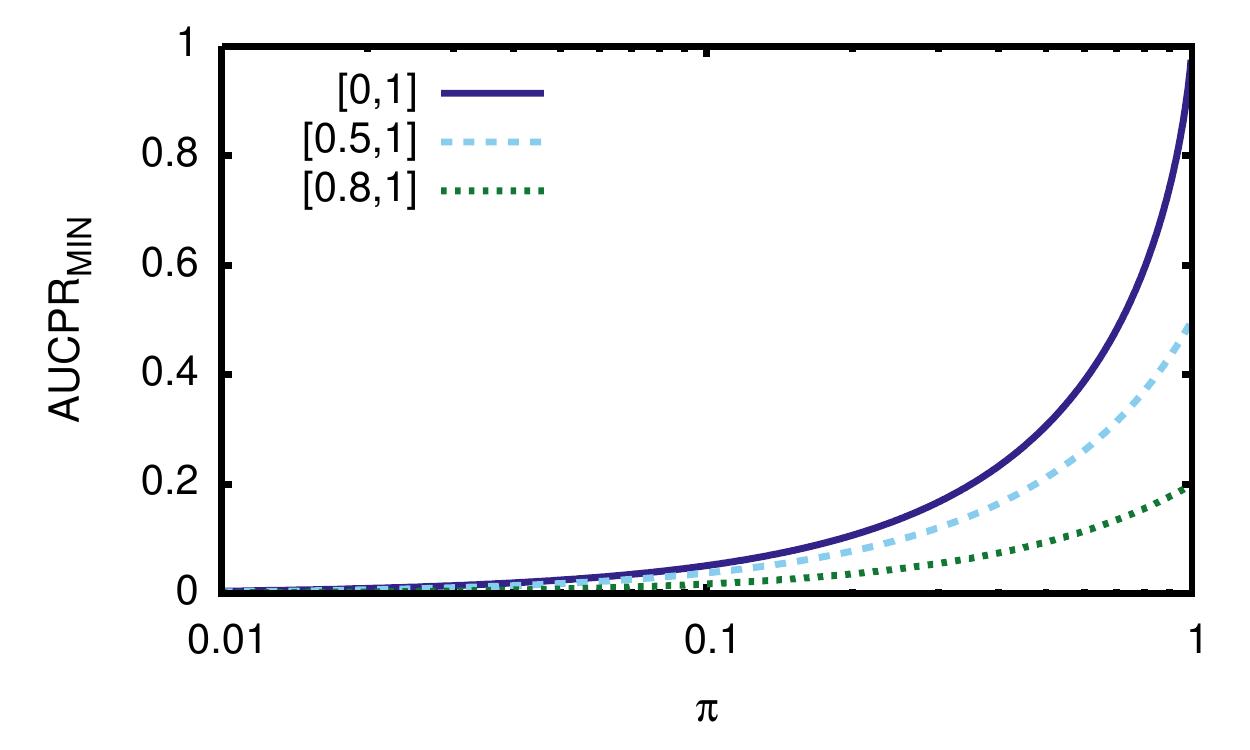}

\begin{samepage}
\begin{corollary} \label{thm:minarearestricted} For calculation of\aucpr over recalls in $[a,b]$ where $0 \leq a < b \leq 1$,
  \begin{equation*} \aucprmin = b - a + \frac{1 - \pi}{\pi} \ln
\left (\frac{\pi(a-1)+1}{\pi(b-1)+1} \right )
  \end{equation*}
\end{corollary}
\end{samepage}

\section{PR Space Metrics that Account for Unachievable Region}

The unachievable region represents a lower bound on AUCPR and it is
important to develop evaluation metrics that account for this.  We
believe that any metric $\mathrm{A'}$ that replaces AUCPR should satisfy at
least the following two properties.  First, $\mathrm{A'}$ should relate to AUCPR.
Assume AUCPR was used to estimate the performance of
classifiers $C_1,\ldots, C_n$ on a \emph{single} test set. If
$\aucpr(C_i,\mathrm{test_D}) > \aucpr(C_j,\mathrm{test_D})$, then $\mathrm{A'}(C_i,\mathrm{test_D}) >
\mathrm{A'}(C_j,\mathrm{test_D})$, as test set $\mathrm{test_D}$'s skew affects each model
equally.  Note that this property may not be appropriate or desirable
when aggregating scores across \emph{multiple} test sets, as done in cross
validation, because each test set may have a different skew.  Second,
$\mathrm{A'}$ should have the same range for every data set,
regardless of skew. This is necessary, though not
sufficient, to achieve meaningful comparisons across data sets. AUCPR
does not satisfy the second requirement because, as shown in
Theorem~\ref{thm:minarea}, its range depends on the data set's skew.

% \aucnpr macro puts spaces after ( and before ) so hard code in
We propose the normalized area under the PR curve ($\mathrm{AUCNPR}$). From AUCPR, we subtract the minimum AUCPR, so the worst ranking has a score of 0. We then normalize so the best ranking has a score of 1.
\begin{equation*} \label{eq:aucnpr}
  \aucnpr = \frac{\aucpr - \aucprmin}{\aucprmax-\aucprmin}
\end{equation*}
where $\aucprmax=1$ when calculating area under the entire PR curve and $\aucprmax = b-a$ when restricting recall to $a \leq r \leq b$.

Regardless of skew, the best possible classifier will have an AUCNPR of 1 and
the worst possible classifier will have an AUCNPR of 0. AUCNPR also preserves the ordering of algorithms on the same
test set since\aucprmax and\aucprmin are constant for the
same data set.  Thus, AUCNPR satisfies our proposed requirements for a replacement of AUCPR.  Furthermore, by
accounting for the unachievable region, it makes comparisons between
data sets with different skews more meaningful than AUCPR.

An alternative to AUCNPR would be to
normalize based on the AUCPR for random guessing, which is simply
$\pi$.  This has two drawbacks.  First, the range of
scores depends on the skew, and therefore is not consistent across
different data sets.  Second, it can result in a negative score if an
algorithm performs worse than random guessing, which seems
counter-intuitive for an area \emph{under} a curve.

A discussion of degenerate data sets with $\pi=0$ or $\pi=1$, where\aucprmin and\aucnpr are undefined, is in our technical report \cite{boyd2012}.

\section{Discussion and Recommendations}
We believe all practitioners using evaluation scores based on PR space (e.g., PR curves, AUCPR, AP, F1) should be cognizant of the unachievable region and how it may affect their analysis.

Visually inspecting the PR curve or looking at an AUCPR score often
gives an intuitive sense for the quality of an algorithm or difficulty
of a task or data set.  If the skew
is extremely large, the effect of the very small unachievable
region is negligible on PR analysis. However, there are many instances where the skew is
closer to 0.5 and the unachievable area is not insignificant.  With
$\pi=0.1$, $\aucprmin \approx 0.05$, and it increases as $\pi$ approaches 0.5.  AUCPR is used in many 
applications where $\pi > 0.1$
\cite{hu2009,sonnenburg2006,liu2007}.  Thus a general awareness of
the unachievable region and its relationship to skew is important when
casually comparing or inspecting PR curves and AUCPR scores.  A simple
recommendation that will make the unachievable region's impact on
results clear is to {\em always show the minimum PR curve on PR curve
plots\/}.

Next, we discuss several specific situations where the unachievable
region is highly relevant.

\subsection{Aggregation for Cross-Validation}
In cross validation, stratification typically allows different folds
to have similar skews. However, particularly in relational domains,
this is not always the case. In relational domains, stratification
must consider fold membership constraints imposed by links
between objects that, if violated, would bias the results of cross
validation.  For example, consider the bioinformatics task of protein
secondary structure prediction. Putting amino acids from the same
protein in different folds has two drawbacks.  First, it could bias
the results as information about the same protein is in both the train
and test set. Second, it does not properly simulate the ultimate goal
of predicting the structure of entirely novel proteins.  Links between
examples occur in most relational domains, and placing all linked
items in the same fold can lead to substantial variation in the skew
of the folds.  Since the different skews yield different\aucprmin, care must be taken when aggregating results to create a
single summary statistic of an algorithm's performance.

Cross validation assumes that each fold is sampled from the same
underlying distribution.  Even if the skew varies across folds, the
merged data set is the best estimate of the underlying distribution
and thus the overall skew.  Ideally, aggregate descriptions, like a PR
curve or AUCPR, should be calculated on a single, merged data set.
Merging directly compares probability estimates for examples in different
folds and assumes that the models are calibrated.  Unfortunately, this is rarely a primary goal of machine learning and learned
models tend to be poorly calibrated \cite{forman2010}.

With uncalibrated models, the most common practice is to average the
results from each fold. For AUCPR, the summary score is the mean of
the AUCPR from each fold. For a PR curve, vertical averaging of the
individual PR curves from each fold provides a summary curve.  In both
cases, averaging fails to account for any difference in the
unachievable regions that arise due to variations in class skew. As
shown in Theorem~\ref{thm:minarea}, the range of possible AUCPR values
varies according to a fold's skew. Similarly, when vertically
averaging PR curves, a particular recall level will have varying
ranges of potential precision values for each fold if the folds have
different skews.
Even a single fold, which has much higher precision values due to a substantially lower skew, can cause a higher vertically averaged PR curve because of its larger unachievable region.
Failing to account for fold-by-fold variation in
skew can lead to overly optimistic assessments when using
straight-forward averaging.

We recommend averaging AUCNPR instead of AUCPR when evaluating area
under the curve.  Averaging AUCNPR, which has the same range
regardless of skew, helps reduce (but not eliminate) skew's effect
compared to averaging AUCPR.  A similar normalization approach for
summarizing the PR curve leads to a non-linear transformation of PR
space that can change the area under the curves in unexpected ways.
An effective method for generating a summary PR curve that preserves
measures of area in a satisfying way and accounts for the unachievable
region would be useful and is a promising area of future
research.

\subsection{Aggregation among Different Tasks}

Machine learning algorithms are commonly evaluated on several
different tasks. This setting differs from cross-validation because
each task is not assumed to have the same underlying distribution.
While the tasks may be unrelated \cite{tang2009}, often they come from the same domain. For
example, the tasks could be the truth values of different predicates
in a relational domain \cite{kok2010,mihalkova2007} or
different queries in an IR setting \cite{manning2008}.
Often, researchers report a single, aggregate score by averaging the
results across the different tasks. However, the tasks can potentially
have very different skews, and hence different minimum
AUCPR. Therefore, averaging AUCNPR scores, which
(somewhat) control for skew, is preferred to averaging
AUCPR.

In SRL, researchers frequently evaluate
algorithms by reporting the average AUCPR over a variety of tasks in a
single data set \cite{mihalkova2007,kok2010}.  As a case study,
consider the commonly used IMDB data set. Here, the task is to predict
the probability that each possible grounding of each predicate is
true. Across all predicates in IMDB the skew of true groundings is
relatively low ($\pi=0.06$), but there is significant variation in the
skew of individual predicates.  For example, the $\tt{gender}$
predicate has a skew close to $\pi = 0.5$, whereas a predicate such as
$\tt{genre}$ has a skew closer to $\pi = 0.05$. While presenting the
mean AUCPR across all predicates is a good first approach, it leads to
averaging values that do not all have the same range.  For example,
the $\tt{gender}$ predicate's range is $[0.31,1.0]$ while the
$\tt{genre}$ predicate's range is $[0.02,1.0]$.  Thus, an AUCPR of 0.4
means very different things on these two predicates.  For the
$\tt{gender}$ predicate, this score is worse than random guessing,
while for the $\tt{genre}$ predicate this is a reasonably high
score. In a sense, all AUCPR scores of 0.4 are not created equal, but
averaging the AUCPR treats them as equals.

Table~\ref{tbl:UWCSE_task_aggregation} shows AUCPR and AUCNPR for each
predicate on a Markov logic network model learned by the LSM algorithm
\cite{kok2010}. Notice the wide range of scores and that AUCNPR gives
a more conservative overall estimate.  AUCNPR is still sensitive
to skew, so an AUCNPR of 0.4 in the aforementioned predicates still does not imply completely comparable performances, but it is closer
than AUCPR.

\maketable{tbl:UWCSE_task_aggregation}
{ % caption
Average AUCPR and AUCNPR scores for each predicate in the IMDB set. Results are for the LSM algorithm from Kok and Domingos \yrcite{kok2010}.
The range of scores shows the difficulty and skews of the prediction tasks vary greatly. 
 By accounting for the (potentially large) unachievable regions, AUCNPR yields a more conservative overall estimate of performance. 
}
{|c|c|c|}
{
    \hline
    Predicate & AUCPR & AUCNPR \\ \hline
    $\tt{actor}$ & 1.000 & 1.000 \\ \hline
    $\tt{director}$ & 1.000 & 1.000 \\ \hline
    $\tt{gender}$ & 0.509 & 0.325 \\ \hline
    $\tt{genre}$ & 0.624 & 0.611 \\ \hline
    $\tt{movie}$ & 0.267 & 0.141 \\ \hline
    $\tt{workedUnder}$ & 1.000 & 1.000 \\ \hline
    mean & 0.733 & 0.680 \\ \hline
}

\subsection{Downsampling}

Downsampling is common when learning on highly skewed tasks.  Often
the downsampling alters the skew on the train set (e.g., subsampling the negatives to facilitate learning, or using data from case-control studies) such that it does
not reflect the true skew.  PR analysis is frequently used on the
downsampled data sets \cite{sonnenburg2006,natarajan2011,ilya2010}.  The sensitivity of AUCPR and
related scores makes it important to recognize,
and if possible quantify, the effect of downsampling on evaluation metrics.

The varying size of the unachievable region provides an explanation
 and quantification of some of the dependence of PR curves and AUCPR
 on skew.  Thus, AUCNPR, which adjusts for the unachievable region,
 should be more stable than AUCPR to changes in skew.  To explore
 this, we used SAYU~\cite{davis2005} to learn a model for the $\tt{advisedBy}$
 task in the UW-CSE domain for several downsampled train sets.
 Table~\ref{tbl:uwcse_downsampling} shows the AUCPR and AUCNPR scores
 on a test set downsampled to the same skew as the train set and on the original (i.e., non-downsampled) test set.  AUCNPR has less
 variance than AUCPR. However, there is still a sizable difference
 between the scores on the downsampled test set and the original test
 set. As expected, the difference increases as the ratio approaches 1 positive to 1 negative. At this ratio, even the AUCNPR score on the downsampled
 data is more than twice the score on the original skew.  This is a
 massive difference and it is disconcerting that it occurs simply by
 changing the data set skew. An intriguing area for future
 research is to investigate scoring metrics that either are less
 sensitive to skew or permit simple and accurate transformations that
 facilitate comparisons between different skews.

\maketable{tbl:uwcse_downsampling}
{% caption
    AUCPR and AUCNPR scores for SAYU on UW-CSE $\tt{advisedBy}$ task for different train set skews. The downsampled columns report scores on a test set with the same downsampled skew as the train set.
    The original skew columns report scores on the original test set with a ratio of 1 positive to 24 negatives ($\pi = 0.04$).
}
{|c||c|c||c|c|}
{  \hline
   & \multicolumn{2}{c||}{Downsampled} & \multicolumn{2}{c|}{Original Skew} \\ \hline
   % used scriptsize to get this table in to fit in the column
   Ratio & \scriptsize{AUCPR} & \scriptsize{AUCNPR} & \scriptsize{AUCPR} & \scriptsize{AUCNPR} \\ \hline
   1:1 & 0.851 & 0.785 & 0.330 & 0.316 \\ \hline
   1:2 & 0.740 & 0.680 & 0.329 & 0.315 \\ \hline
   1:3 & 0.678 & 0.627 & 0.343 & 0.329 \\ \hline
   1:4 & 0.701 & 0.665 & 0.314 & 0.299 \\ \hline
   1:5 & 0.599 & 0.560 & 0.334 & 0.320 \\ \hline
   1:10& 0.383 & 0.352 & 0.258 & 0.242 \\ \hline
   1:24& 0.363 & 0.349 & 0.363 & 0.349 \\ \hline 
}

\subsection{F1 Score}
A commonly used evaluation metric for a single point in PR space is the $F_\beta$ family,
\begin{equation*}
  F_\beta = \frac{(1+\beta^2) p r}{\beta^2 p + r}
\end{equation*}
where $\beta > 0$ is a
parameter to control the relative importance of recall and precision
\cite{manning2008}.  
Most frequently, the F1 score ($\beta=1$), which is the harmonic mean of
precision and recall, is used.  We focus our discussion on the F1
score, but similar analysis applies to $F_\beta$.  Figure~\ref{fig:f1_contours} shows contours of the F1 score over PR space.

While the unachievable region of PR space does not put any bounds on
F1 score based on skew, there is still a subtle interaction between skew and
F1.  Since F1 combines precision and recall into a single
score, it necessarily loses information. One aspect of this information loss is that PR points with the same F1 score can have vastly different relationships with the unachievable region.  Consider points A, B, and C in
Figure~\ref{fig:f1_contours}. All three have an F1 score of 0.45, but
each has a very different interpretation if obtained from a data set
with $\pi = 0.33$.  Point A is unachievable and no valid confusion
matrix for it exists. Point B is achievable, but is very near the
minimum PR curve and is only marginally better than random
guessing. Point C has reasonable performance representing good
precision at modest recall.

While losing information is inevitable with a summary like F1, the different interpretations arise partly because F1 treats recall and precision interchangeably.
Furthermore, this is not unique to $\beta=1$. 
While $F_{\beta}$ changes the relative importance, the assumption remains that precision and recall, appropriately scaled by $\beta$, are equivalent for assessing performance.
Our results on the unachievable region show this is problematic as recall and precision have fundamentally
different properties. Every recall has a minimum precision, while there
is a maximum recall for low precision, and no constraints for most
levels of precision.

While a modified F1 score that is sensitive to the unachievable region would be useful, initial work suggests an ideal solution may not exist. Consider three simple requirements for a modified F1 score, $f'$:
\begin{align}
  f'(r,p) = 0 \ \text{if} \ p = \frac{r \pi}{1 - \pi + r\pi} \label{eq:prop_mincurve}\\
  f'(r_1,p) < f'(r_2,p) \ \text{iff} \ r_1 < r_2 \label{eq:prop_recall}\\
  f'(r,p_1) < f'(r,p_2) \ \text{iff} \ p_1 < p_2 \label{eq:prop_precision}
\end{align}
\myeqref{eq:prop_mincurve} ensures $f'=0$ if the PR point is on the minimum PR curve
and Eqs.~\eqref{eq:prop_recall} and \eqref{eq:prop_precision} capture the expectation that an increase in
precision or recall while the other is constant should always increase $f'$.  However, these three properties are impossible to
satisfy because they require $0 = f'(0,0) < f'(0,\pi) < f'(1,\pi) =
0$.  Relaxing Eqs.~\eqref{eq:prop_recall} and \eqref{eq:prop_precision} to $\leq$ makes it
possible to construct an $f'$ that satisfies the requirements but implies $f'(r,p) = 0$ if $p \leq \pi$.  This seems unsatisfactory because it ignores all distinctions once the performance is worse than random guessing. One modified F1 score that satisfies the relaxed
requirements would assign 0 to any PR point worse than random guessing and use the harmonic mean of recall and $\frac{p-\pi}{1-\pi}$ (precision normalized to random guessing) otherwise.

Ultimately, while F1 score or a modified F1 score can be extremely useful, nuanced analyses must never overlook that it is a summary metric, and vital information for interpreting a model's performance may be lost in the summarizing.

\makefigure{fig:f1_contours}{Contours of F1 score in PR space with the minimum PR curve and unachievable region for $\pi = 0.33$. The points A, B, and C all have $\text{F1}=0.45$, but lead to substantially different practical interpretations.}{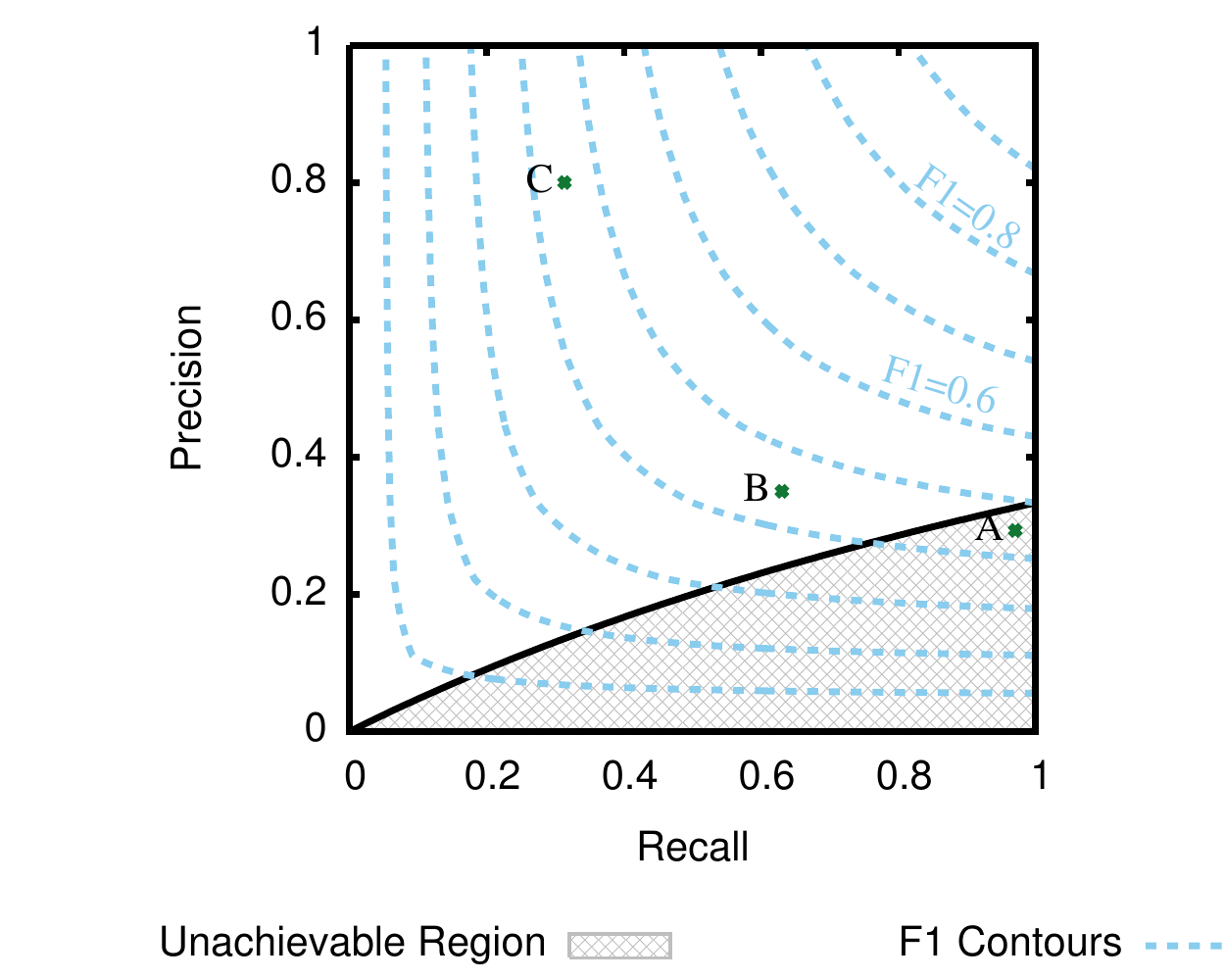}

\section{Conclusion}
We demonstrate that a region of precision-recall space is unachievable for any particular ratio of positive to negative examples.
With the precise characterization of this unachievable region given in Theorems~\ref{thm:constraint}~and~\ref{thm:minarea}, we further the understanding of the effects of downsampling and the impact of the minimum PR curve on F measure and score aggregation.

\begin{small}
\section*{Acknowledgments}
We thank Jude Shavlik and the anonymous reviewers for their insightful
comments and suggestions, and Stanley Kok for providing the LSM algorithm results. We gratefully acknowledge our funding
support.  KB is funded by NIH 5T15LM007359.  VC by the ERDF through
the Progr.\ COMPETE, the Portuguese Gov.\ through FCT, proj.\ HORUS
ref.\ PTDC/EIA-EIA/100897/2008, and the EU Sev.\ Fram.\ Progr.\
FP7/2007-2013 under grant aggrn 288147.  JD by the Research Fund
K.U. Leuven (CREA/11/015 and OT/11/051), EU FP7 Marie Curie Career
Integration Grant (\#294068), and FWO-Vlaanderen (G.0356.12).  DP by
NIGMS grant R01GM097618-01, NLM grant R01LM011028-01, NIEHS grant
5R01ES017400-03 and the UW Carbone Cancer Center.

\end{small}

\begin{small}
\bibliographystyle{icml2012}
\bibliography{prspace}
\end{small}

\end{document}